\documentclass[10pt,journal]{IEEEtran}
\usepackage{enumitem,calc}
\usepackage{graphicx}
\usepackage{subfigure}
\usepackage{amsmath}
\usepackage{array}
\usepackage{subfig}
\usepackage{tabu}
\usepackage{amssymb}
\usepackage{multirow}
\usepackage{booktabs}
\usepackage{cite}
\usepackage{soul}
\soulregister\cite7
\soulregister\mathbf7

\usepackage[table]{xcolor}

\usepackage{rotating}
\usepackage{array}
\usepackage{makecell}

\ifCLASSINFOpdf
\else
\fi
\begin{document}

\title{Expertise-aware Multi-LLM Recruitment and Collaboration for Medical Decision-Making}

\author{
        Liuxin Bao*,
        Zhihao Peng*,
        Xiaofei Zhou\textsuperscript{{\dag}},
        Runmin Cong,
        Jiyong~Zhang, 
        Yixuan Yuan\textsuperscript{{\dag}}

\thanks{* These authors contributed equally to this work.}
\thanks{{\dag} Corresponding author.}

\thanks{L. Bao, X. Zhou, and J. Zhang are with the School of Automation, Hangzhou Dianzi University, Hangzhou 310018, China (e-mail: lxbao@hdu.edu.cn; zxforchid@outlook.com; jzhang@hdu.edu.cn).}

\thanks{R. Cong is with the School of Control Science and Engineering, Shandong University, Jinan 250101, China (e-mail: rmcong@sdu.edu.cn).}

\thanks{Z. Peng and Y. Yuan are with the Department of Electronic Engineering, Chinese University of Hong Kong, Shatin 000000, China (e-mail: zhihao.peng@cityu.edu.hk; yxyuan@ee.cuhk.edu.hk).}




}

\maketitle

\begin{abstract}
Medical Decision-Making (MDM) is a complex process requiring substantial domain-specific expertise to effectively synthesize heterogeneous and complicated clinical information. While recent advancements in Large Language Models (LLMs) show promise in supporting MDM, single-LLM approaches are limited by their parametric knowledge constraints and static training corpora, failing to robustly integrate the clinical information. To address this challenge, we propose the Expertise-aware Multi-LLM Recruitment and Collaboration (EMRC) framework to enhance the accuracy and reliability of MDM systems. It operates in two stages: ($i$) expertise-aware agent recruitment and ($ii$) confidence- and adversarial-driven multi-agent collaboration. 
Specifically, in the first stage, we use a publicly available corpus to construct an LLM expertise table for capturing expertise-specific strengths of multiple LLMs across medical department categories and query difficulty levels. This table enables the subsequent dynamic selection of the optimal LLMs to act as medical expert agents for each medical query during the inference phase. 
In the second stage, we employ selected agents to generate responses with self-assessed confidence scores, which are then integrated through the confidence fusion and adversarial validation to improve diagnostic reliability. 
We evaluate our EMRC framework on three public MDM datasets, where the results demonstrate that our EMRC outperforms state-of-the-art single- and multi-LLM methods, achieving superior diagnostic performance. For instance, on the MMLU-Pro-Health dataset, our EMRC achieves 74.45\% accuracy, representing a 2.69\% improvement over the best-performing closed-source model GPT-4-0613, which demonstrates the effectiveness of our expertise-aware agent recruitment strategy and the agent complementarity in leveraging each LLM's specialized capabilities. Our code is available at https://github.com/Lx-Bao/EMRC.
\end{abstract}



\IEEEpeerreviewmaketitle

\section{Introduction}
\label{sec:introduction}

Medical Decision-Making (MDM) is a clinically critical and intellectually rigorous process that underpins effective clinical practice, requiring physicians to synthesize diverse and heterogeneous sources of clinical information to arrive at accurate, context-sensitive diagnostic and therapeutic conclusions \cite{masic2022medical, corrao2022rethinking}. This process demands a high level of expertise-specific knowledge, as it integrates empirical evidence with specialized medical knowledge to address complex patient cases under conditions of uncertainty and stringent time constraints \cite{croskerry2018adaptive, 9627588}. The effective MDM system fundamentally relies on interprofessional collaboration, where domain-specific expertise is critical for resolving diagnostic uncertainties and delivering reliable patient-centric care. The inherent complexity of MDM necessitates leveraging the specialized knowledge to ensure the clinical accuracy and reliability \cite{bujold2022decision,10173552,10697408}.

In recent years, the rapid evolution of Large Language Models (LLMs) \cite{ouyang2022training,achiam2023gpt,chen2024more,liang2024can,yuksekgonul2025optimizing,wang2025towards, 10854911} has catalyzed a surge of interest in their potential to support MDM. Leveraging their capacity for natural language understanding, contextual reasoning, and evidence retrieval, researchers have begun investigating the integration of LLMs into clinical workflows to support and enhance the MDM capability of physicians \cite{tang2023medagents, ke2024mitigating, wei2024medco, kim2024mdagents, wang2025colacare,yue2024clinicalagent,li2024agent,hong2024argmed}. For example, 
Tang \emph{et al.} \cite{tang2023medagents} enabled zero-shot medical reasoning by orchestrating LLMs in role-based collaborative discussions through expert gathering, analysis, summarization, consultation, and decision-making stages.
Kim \emph{et al.} \cite{kim2024mdagents} proposed an adaptive framework that dynamically organizes LLMs into solo or collaborative teams based on the complexity of queries to emulate the real-world MDM. 
Hong \emph{et al.} \cite{hong2024argmed} empowered LLMs to perform an explainable MDM by simulating clinical discussions through iterative argument generation and verification based on formal argumentation schemes, culminating in decision support via symbolic reasoning. 

\begin{figure}[!t]
  \centering
\begin{tabular}{c@{\hspace{1pt}}c}
     \includegraphics[width = 0.3\textwidth]{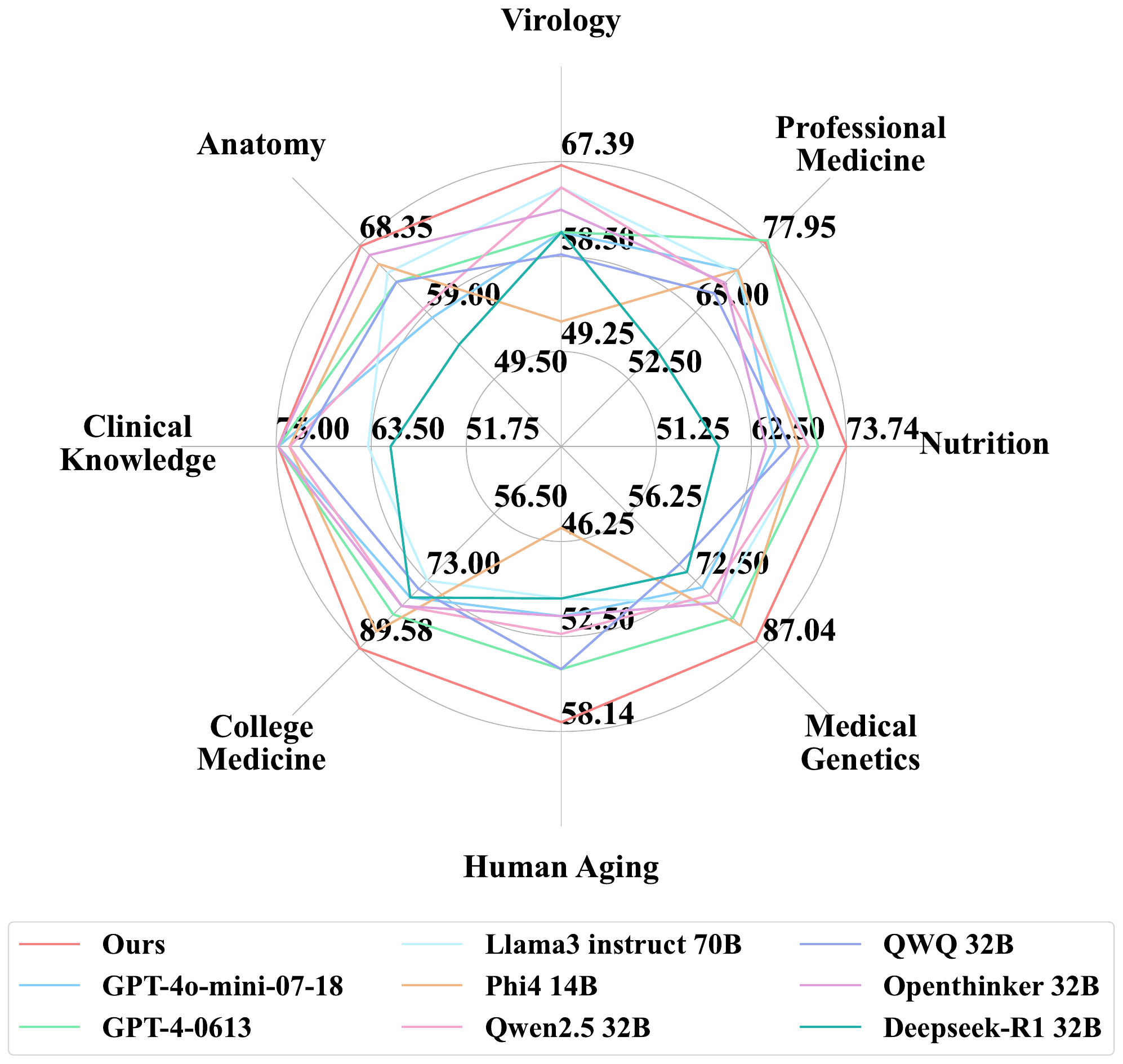}&
     \includegraphics[width = 0.18\textwidth]{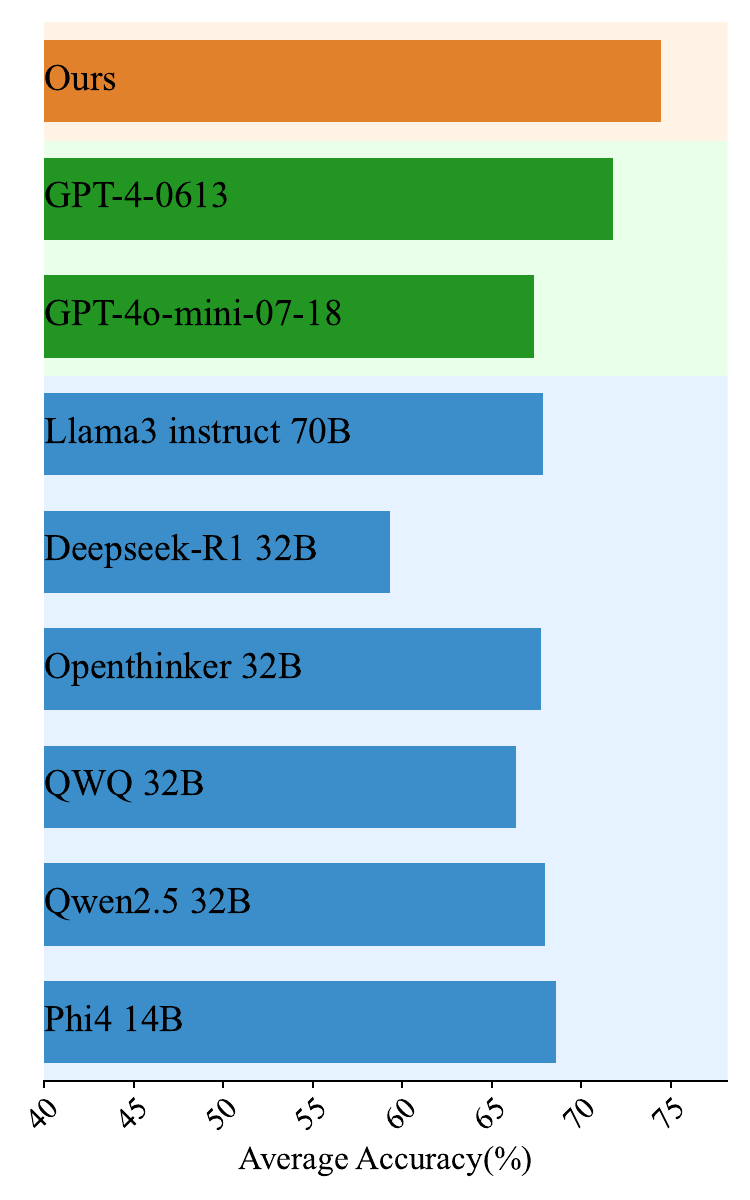} 
\end{tabular}
\caption{The performance of different LLMs across various medical domains with specialized expertise on the MMLU-Pro-Health dataset.}
\label{fig_perform_MP}
\end{figure}



\begin{figure*}
\centering
\includegraphics[width=0.99\textwidth]{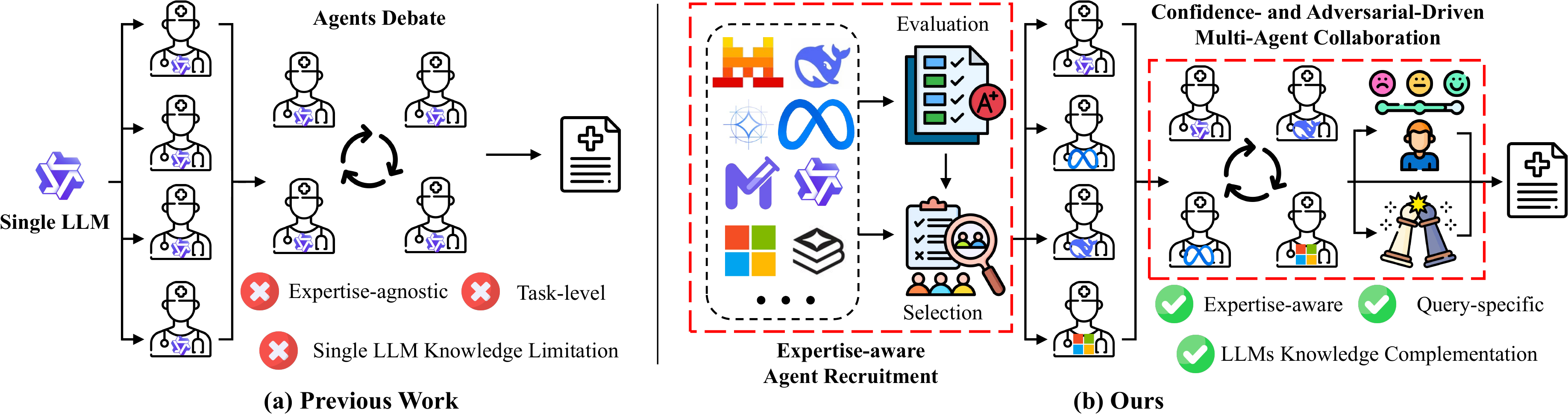}
\caption{{Architecture comparison between previous work and Ours.}}
\label{fig_Intro}
\end{figure*}

Despite the progress made by existing MDM studies \cite{tang2023medagents, ke2024mitigating, wei2024medco, kim2024mdagents, wang2025colacare,yue2024clinicalagent,li2024agent,hong2024argmed}, the application of LLM collaboration in support of MDM still presents two key challenges due to the inherent complexity of MDM: 

\textbf{(\textit{i}) The requirements of the diverse and specialized knowledge across different medical domains.}  
To cope with this challenge, existing MDM frameworks adopt a role-playing paradigm \cite{tang2023medagents,kim2024mdagents,li2024agent}, in which multiple agents with domain-specific roles are instantiated from a single LLM to tackle queries spanning various medical domains, as illustrated in Fig.~\ref{fig_Intro}(a). However, such frameworks typically rely on a single specific LLM and adopt an expertise-agnostic selection strategy that treats all LLMs as functionally equivalent and overlooks their specialized strengths in different medical domains. Meanwhile, the same LLM is uniformly applied to every query in the task level, without accounting for the differences of each query in domain specificity or clinical complexity. Consequently, these frameworks are restricted to the limitations of a single model's knowledge base, which may be incomplete, outdated, or insufficient for certain domains, thereby constraining the diversity and currency of information critical for accurate MDM.
Furthermore, as shown in Fig.~\ref{fig_perform_MP}, our empirical results reveal that although several LLMs achieve comparable overall accuracy on the MMLU-Pro-Health dataset, their performance varies considerably across distinct categories of medical queries. Such variability arises from differences in model architecture, training corpora, and optimization objectives \cite{wang2022language, hadi2023survey}, leading to divergent representation capabilities and reasoning patterns. These findings underscore a critical insight: \textit{no single LLM consistently outperforms others across all medical domains when models operate at comparable scales}, paralleling the reality of clinical practice where no individual expert possesses exhaustive knowledge across all specialties. Moreover, the results in Fig.~\ref{fig_perform_MP} further indicate that each LLM demonstrates expertise-specific strengths, highlighting the potential of integrating complementary capabilities to enhance the accuracy and reliability of MDM. Therefore, for each input query, an ideal MDM framework should adaptively select a subset of LLMs with the highest domain-specific expertise to fulfill its distinct knowledge demands.

\textbf{(\textit{ii}) The difficulty in ensuring consistency and accuracy when integrating information from multiple sources.}
MDM tasks inherently involve the integration of complex and diverse information from multiple sources (\textit{e.g.,} diagnosis from different physicians). Existing MDM frameworks often rely on straightforward collaboration strategies, such as majority voting \cite{yang2023one}, sequential integration \cite{kim2024mdagents}, or multi-round debate \cite{tang2023medagents}, to aggregate outputs from multiple agents. However, these approaches overlook the varying reliability of individual agents' responses across different medical domains and miss opportunities for iterative refinement through inter-agent feedback. As a result, the outputs of these frameworks are plagued by inconsistent response quality, inadequate validation of reasoning chains, and compromised diagnostic credibility. Moreover, such frameworks are prone to errors or biases, as they fail to systematically resolve contradictions between the agents' contributions, ultimately leading to erosion of overall framework performance.

To address these two aforementioned challenges in MDM, we propose the Expertise-aware Multi-LLM Recruitment and Collaboration (EMRC) framework (Fig.~\ref{fig_Intro}(b)). This framework operates through two sequentially integrated stages: expertise-aware agent recruitment and confidence- and adversarial-driven multi-agent collaboration, designed to enhance diagnostic accuracy and reliability by leveraging the complementary strengths of multiple LLMs.
Specifically, the first stage aims to dynamically tailor LLM recruitment to the diverse and specialized knowledge demands of individual medical queries. Here, we construct an LLM expertise table, which systematically quantifies the domain-specific strengths of each model, facilitating the optimal recruitment of agents with demonstrated proficiency in pertinent medical departments and query difficulty levels. 
The second stage seeks to improve the reliability of Medical Decision-Making (MDM) outputs by integrating and refining the responses provided by the selected agents. It is structured within a multi-layer architecture that prioritizes high-quality contributions while systematically addressing inconsistencies. During this stage, the agents generate responses accompanied by self-assessed confidence scores, which are subsequently refined through both confidence fusion and adversarial validation. This iterative process of refinement aids in the identification and resolution of discrepancies, thereby strengthening the overall reliability of the decision-making process.

In summary, the main contributions of this paper can be summarized as follows:\begin{itemize}[leftmargin=*]

\item We propose a novel EMRC framework, which aims to dynamically recruit agents with expertise-specific knowledge tailored to each medical query, while achieving efficient collaboration between the agents. By leveraging the complementary knowledge across multiple LLMs, our EMRC enhances the accuracy and reliability of MDM.

\item We propose an expertise-aware agent recruitment strategy that pre-assesses prior domain expertise of individual LLMs and dynamically leverages this information to recruit optimally specialized LLMs as medical expert agents during inference.

\item We propose a confidence- and adversarial-driven multi-agent collaboration strategy that capitalizes on the complementary strengths of multiple agents by implementing a confidence fusion mechanism and an adversarial validation process. 

\item Extensive experiments on three public MDM datasets demonstrate that the EMRC framework outperforms state-of-the-art single- and multi-LLM methods, achieving superior diagnostic accuracy.




\end{itemize}


\section{Related Work}
\subsection{LLM Reasoning}
LLMs have demonstrated remarkable performance improvements across a range of areas, including mathematics, science, and programming \cite{wei2022chain,zhou2022least,yao2023tree,besta2024graph, 10970423,11015274,11086426,10543121}. These improvements are largely driven by advances in reasoning strategies that substantially enhance the models’ ability to solve complex problems.
Among these strategies, Chain-of-Thought (CoT) guided models to generate intermediate reasoning steps, thereby improving performance on multi-step tasks \cite{wei2022chain}. The Least-to-Most (LtM) further advanced this approach by decomposing a task into a sequence of subproblems, where the solution to each subproblem provided context for the next \cite{zhou2022least}. The Tree-of-Thought (ToT) introduced a tree-based structure, enabling the model to explore multiple reasoning paths in parallel \cite{yao2023tree}. The Skeleton-of-Thought (SoT) enhanced generation efficiency by first producing a high-level outline of the response and then filling in the details in parallel \cite{ning2023skeleton}. More recently, Graph-of-Thought (GoT) proposed a dynamic reasoning paradigm that represented the reasoning process as a graph of interconnected thought nodes, allowing flexible and context-sensitive exploration \cite{besta2024graph}.

Recent studies have further extended the reasoning capabilities of LLMs, where ongoing efforts focus on enhancing multi-step reasoning, factual consistency, and domain-specific inference capabilities. For example, natural language embedding programs enhanced model performance in mathematical and symbolic reasoning tasks by integrating natural language with programmatic structures \cite{zhang2023natural}. In addition, models such as Deepseek-R1 \cite{guo2025deepseek}, optimized through reinforcement learning, exhibited substantial improvements in reasoning tasks, achieving performance comparable to state-of-the-art models. These advancements indicated that the reasoning abilities of LLMs were continuously strengthened through innovative techniques and model optimization.

\subsection{Multi-agent Collaboration}
Recent research has demonstrated that strategically combining complementary strengths across multiple LLMs through collaboration frameworks could significantly enhance performance on complex, knowledge-intensive tasks \cite{li2025rethinking, kim2024mdagents, wang2025mixtureofagents}. Broadly, existing collaboration frameworks could be categorized into two principal paradigms: role-playing and multi-LLM debate.
In the role-playing framework, individual LLMs were assigned specialized roles or responsibilities, each concentrating on subtasks aligned with their designated functions \cite{kim2024mdagents, wang2024survey}. Through structured cooperative interactions, these models collectively pursued and accomplished complex, high-level goals. This paradigm not only enabled systematic problem decomposition via explicit division of labor, but also capitalized on each model’s specialized competencies to synthesize coherent and integrated solutions.
In contrast, the multi-LLM debate framework adopted a more decentralized strategy, wherein each model independently formulated a solution and subsequently engaged in critical evaluation of the others’ outputs to iteratively refine and converge on a consensus \cite{du2023improving, chen2023reconcile}. Depending on the configuration of participating models and the nature of their interaction protocols, this framework could be further divided into several nuanced subtypes (\emph{e.g.,} self-consistency voting \cite{wang2022self} and critique-and-revise \cite{shinn2023reflexion}), each with distinct mechanisms of coordination and resolution.
Beyond these dominant approaches, a variety of alternative collaborative paradigms have also been explored. Social learning, for example, facilitated model improvement through observational learning, enabling LLMs to adapt their internal decision-making policies or generation strategies based on the behaviors or outputs of peers \cite{mohtashami2023social}. Similarly, the chain-of-agents framework enhanced performance on long-context or multi-step reasoning tasks by organizing models in structured, sequential workflows that promoted inter-agent dependency and coordination \cite{zhang2024chain}. Additionally, recent surveys examining collaboration between large and small models have proposed a spectrum of interaction strategies, including pipelining, routing, assistance, distillation, and fusion, which further diversified the collaborative landscape \cite{wang2025mixtureofagents, chen2025symbolic, fu2023improving}.

Collectively, these studies have reflected an increasingly sophisticated understanding of LLM collaboration, offering critical insights into the design of robust, scalable, and adaptive multi-agent artificial intelligence systems.

\begin{figure*}
\centering
\includegraphics[width=0.99\textwidth]{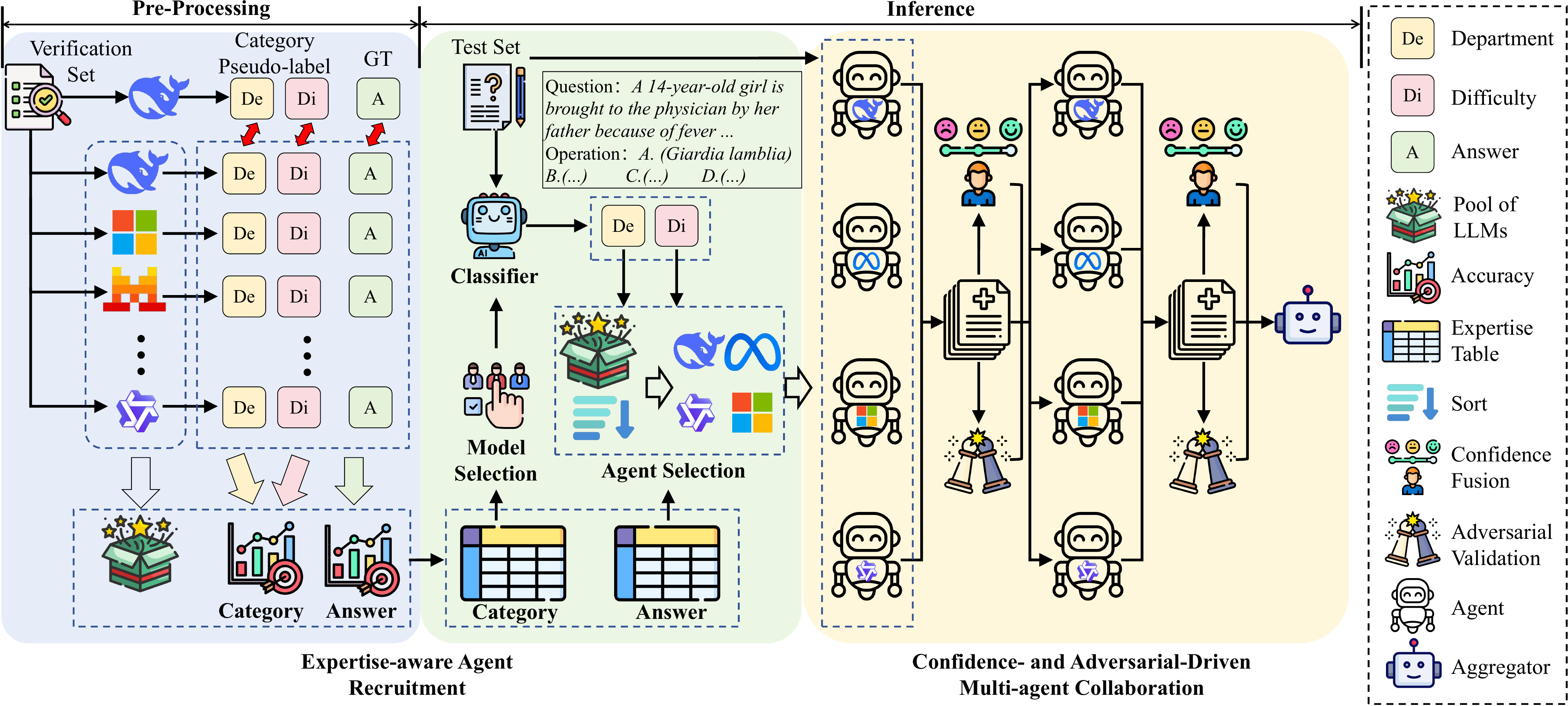}
\caption{{The architecture of the proposed EMRC framework.}The framework consists of two main stages: (1) expertise-aware agent recruitment, where candidate LLMs are evaluated to construct an LLMs expertise table and dynamically selected based on query category. (2) Confidence- and adversarial-driven multi-agent collaboration, where selected agents iteratively refine responses through confidence fusion and adversarial verification.}
\label{fig_overall}
\end{figure*}

\subsection{Medical Reasoning Agent}
In MDM, medical LLMs have shown promising capabilities in automating tasks such as medical question answering and diagnostic reasoning. For example, \textit{Meditron} \cite{chen2023meditron} focused on optimizing medical language models by incorporating domain-specific pretraining on a carefully curated corpus, which included the use of medical guidelines, public medical articles, abstracts, and experience replay data to improve medical reasoning capabilities. \textit{HuatuoGPT} \cite{zhang2023huatuogpt} combined supervised fine-tuning with hybrid data and reinforcement learning with AI feedback, which trained on both ChatGPT’s distilled data and real-world data from physicians to improve diagnostic accuracy and response quality.

However, standalone LLMs faced significant challenges in executing complex, multi-step clinical reasoning and mitigating hallucination risks, potentially compromising their reliability in high-stakes MDM scenarios. To address these limitations, recent advances in medical reasoning agent frameworks have demonstrated the potential to enhance MDM by mitigating the constraints of standalone LLMs. These systems employed specialized agent roles, including intent interpretation, differential diagnosis generation, and therapeutic strategy formulation—enabling context-aware, patient-specific healthcare delivery. For example, the system developed by Ke \emph{et al.}\cite{ke2024mitigating} reduced cognitive biases in diagnosis by using agents that provided expert opinions and critical evaluations.
\textit{MedAide}\cite{wei2024medco} coordinated agents across stages including pre-diagnosis, diagnosis, medication, and post-diagnosis, while frameworks such as \textit{MDagents}\cite{kim2024mdagents} and \textit{EHRagent}\cite{wang2025colacare} improved diagnostic accuracy through structured discussions and shared reasoning. Domain-specific applications also showed promise, with several frameworks tailoring multi-agent collaboration to specialized tasks. Yue \emph{et al.}\cite{yue2024clinicalagent} used multi-agent collaboration to predict clinical trial outcomes by integrating large-scale domain knowledge. The \textit{Polaris} framework\cite{mukherjee2024polaris} combined general communication agents with task-specific agents to ensure safe patient interactions.
Li \emph{et al.} \cite{li2024agent} constructed a fully simulated hospital environment (\emph{i.e.,} \textit{Agent Hospital}), where LLM-powered physician agents evolved by interacting with synthetic patients generated via coupling LLMs with medical knowledge bases, enabling continuous expertise acquisition without manual annotation and achieving strong real-world generalization.

\section{Proposed Method}
\subsection{Problem Formulation}

Given a medical query $Q$ and a pool of heterogeneous LLMs $\mathcal{M} = \{ M_1, M_2, \dots, M_L \}$, the objective is to build a framework that selects an optimal subset of LLMs for $Q$ and generates a reliable medical diagnosis through multi-agent collaboration:
\begin{equation}
\hat{y}_q = \mathcal{C}\big(Q, \mathcal{S}(Q, \mathcal{M})\big),
\end{equation}
where $\mathcal{S}(\cdot)$ is the selection function that returns the subset $\mathcal{M}_q \subseteq \mathcal{M}$ satisfying the resource constraint $|\mathcal{M}_q| \leq N_{\max}$, $N_{\max}$ is the maximum number of recruited LLMs. $\mathcal{C}(\cdot)$ is the multi-agent collaboration function that integrates the outputs of the selected LLMs to produce the final diagnosis $\hat{y}_q$.

In contrast to conventional single-LLM approaches, our framework dynamically selects models that are most capable of handling each query and coordinates their interaction through confidence fusion and adversarial verification. This design leverages the complementary expertise of multiple LLMs, enhancing both the accuracy and reliability of automated medical decision-making.

\subsection{Overall Architecture}

As illustrated in Fig.~\ref{fig_overall}, the proposed Expertise-aware Multi-LLM Recruitment and Collaboration (EMRC) framework operates through two interconnected stages: expertise-aware agent recruitment and confidence- and adversarial-driven multi-agent collaboration. These stages leverage the complementary strengths of multiple LLMs to enhance the accuracy and reliability of MDM.
In the first stage, candidate LLMs are evaluated on the MedQA validation set~\cite{jin2021disease} to assess their query classification proficiency and diagnostic accuracy across medical departments and difficulty levels. This yields an LLM expertise table documenting domain-specific strengths. During inference, the table enables dynamic selection of the most proficient LLM subset as specialized medical expert agents, based on composite expertise scores.
The second stage promotes reliable collaboration in a multi-layer architecture. Each agent generates a response with a self-assessed confidence score, fused with its expertise score for weighted reliability. The highest-expertise agent serves as a \textit{Judge} for adversarial cross-validation, identifying errors. The outputs, including fused scores, errors, and responses, are iteratively refined in subsequent layers. Finally, an \textit{Aggregator} agent synthesizes these outputs into a unified high-quality response.

\subsection{Expertise-aware Agent Recruitment}
The existing medical reasoning agent frameworks \cite{tang2023medagents,kim2024mdagents,li2024agent} primarily adopt a role-playing approach, in which multiple instances with distinct responsibilities (\emph{e.g.,} patient navigator and specialists from various departments) are generated from a single LLM. However, these frameworks tend to overlook the complementary strengths and diverse knowledge bases of different LLMs. Therefore, we propose an expertise-aware agent recruitment strategy that focuses on leveraging the expertise knowledge of different LLMs across different medical department categories and query difficulty levels. 
Our expertise-aware agent recruitment is structured in two stages.

\subsubsection{LLMs Expertise Evaluation}
Consider the pool of LLMs $\mathcal{M}$ consisting of $n$ distinct LLMs from the open-access community, our objective is to recruit the most appropriate subset $\mathcal{M}_{q}$ of LLMs from $\mathcal{M}$ to address each specific medical query $\mathcal{Q}$. The cornerstone of this approach is the precise evaluation of each LLM's potential to deliver accurate responses to the given query. An intuitive strategy is to categorize the query and select optimally specialized LLMs that perform best within that category \cite{li2025rethinking, chen2025symbolic}. However, existing MDM datasets \cite{jin2021disease,katz2024gpt,wang2024mmlu} differ considerably in their internal classification schemes, lacking a standardized classification criterion. This heterogeneity limits the generalizability of category-based recruitment strategies across different benchmarks.

To address this limitation, we design a unified classification strategy for medical queries, structured along two dimensions: medical department categories and query difficulty levels. For the dimension of the medical department categories, we adopt a standardized classification criterion inspired by clinical practice and healthcare organizational structures. This criterion references established categorizations used in medical education and hospital systems \cite{landefeld2016structure, braunwald2006departments}, and has been validated by medical professionals to ensure a broad coverage of general and specialized medical domains commonly encountered in real-world diagnostic scenarios. Specifically, we group queries into nine major departments: Internal Medicine (IM), Surgery (Su), Obstetrics and Gynecology (OG), Pediatrics (Pe), Neurology (Ne), Oncology (On), Otolaryngology (Ot), Psychiatry and Psychology (PP), and Emergency and Critical Care (EC).
For the dimension of the query difficulty level, we stratify the queries into three levels (\emph{i.e.,} low (L), medium (M), and high (H)) according to their cognitive complexity and reasoning demands. This classification criterion provides a practical and generalizable foundation for multi-LLM recruitment. 
Based on the aforementioned classification criteria, we construct an LLM expertise table, which records the performance of each LLM in the pool across different combinations of medical department categories and query difficulty levels. The table is generated by evaluating each LLM on a validation set and quantifying its classification and answer accuracy within each defined category, thereby capturing the nuanced strengths of each model across a diverse range of medical domains.

Specifically, as illustrated on the blue background of Fig.~\ref{fig_overall}, we first introduce a pseudo-labeling approach powered by an LLM with large parameter size to achieve consistent categorization of queries within the validation set. We employ the Deepseek-R1 model \cite{guo2025deepseek} with 671 billion parameters to systematically classify all validation queries in the MedQA dataset, generating corresponding medical department category labels $Dept_{i}^{v}$ and query difficulty level labels $Diff_{i}^{v}$:
\begin{equation}\label{eq1}
Dept_i^v, \; Diff_i^v = \text{Deepseek-R1}(Q_i^v),
\end{equation}
where $Q_{i}^{v}$ represents the $i^{th}$ queries in the validation set of MedQA dataset. Notably, we adopt the MedQA validation set due to its broad coverage of medical domains and real-world clinical relevance, which is suitable for capturing fine-grained expertise.

Subsequently, each LLM with small parameter size (\emph{i.e.,} 8B - 34B) in the pool of LLMs is tasked with predicting the medical department category $Dept_{i}$, query difficulty level $Diff_{i}$, and answer $Ans_{i}$ for each query:
\begin{equation}\label{eq1}
Dept_i^n, \; Diff_i^n, \; Ans_i^n = \text{LLM}_n(Q_i^v),\ \text{LLM}_n \in \mathcal{M}.
\end{equation}

Based on the category pseudo-labels $\{Dept_i^v, Diff_i^v\}$ and LLMs' responses $\{Dept_i^n,  Diff_i^n\}$, we can compute each LLM's classification accuracy for medical department categories and query difficulty levels. Taking $n^{th}$ LLM as an example, the whole process can be formulated as:
\begin{equation}\label{eq1}
\left \{\begin{array}{l}

\text{Acc}_{\text{dept}}^n = \frac{1}{K} \sum_{i=1}^{K} \mathbb{I}\left[Dept_i^n = Dept_i^v\right], \\

\text{Acc}_{\text{diff}}^n = \frac{1}{K} \sum_{i=1}^{K} \mathbb{I}\left[Diff_i^n = Diff_i^v\right],

\end{array}
\right.
\end{equation}
where $\mathbb{I}$ means the indicator function, which equals $1$ if the condition is true, $0$ otherwise. $K$ is the number of the queries in the validation set.

Meanwhile, we can also acquire the answer accuracy across different medical department categories and query difficulty levels of each LLM, which can be formulated as:
\begin{equation}\label{eq1}
\left \{\begin{array}{l}
\text{Acc}_{\text{ans}}^{n,d} = \frac{1}{|I_d|} \sum_{i \in I_d} \mathbb{I}\left[Ans_i^n = GT_i\right], \quad \forall d \in \mathcal{D}, \\
\text{Acc}_{\text{ans}}^{n,l} = \frac{1}{|I_l|} \sum_{i \in I_l} \mathbb{I}\left[Ans_i^n = GT_i\right], \quad \forall l \in \mathcal{L},
\end{array}
\right.
\end{equation}
where $GT_i$ represents the ground truth answer for the query $Q_i^v$. $\mathcal{D}$ and $\mathcal{L}$ mean sets of all department categories and query difficulty levels, respectively. Here, 
$I_d = \{i \mid Dept_i^v = d\}$ and $I_l = \{i \mid Diff_i^v = l\}$. 

\begin{figure}
\centering
\includegraphics[width=0.48\textwidth]{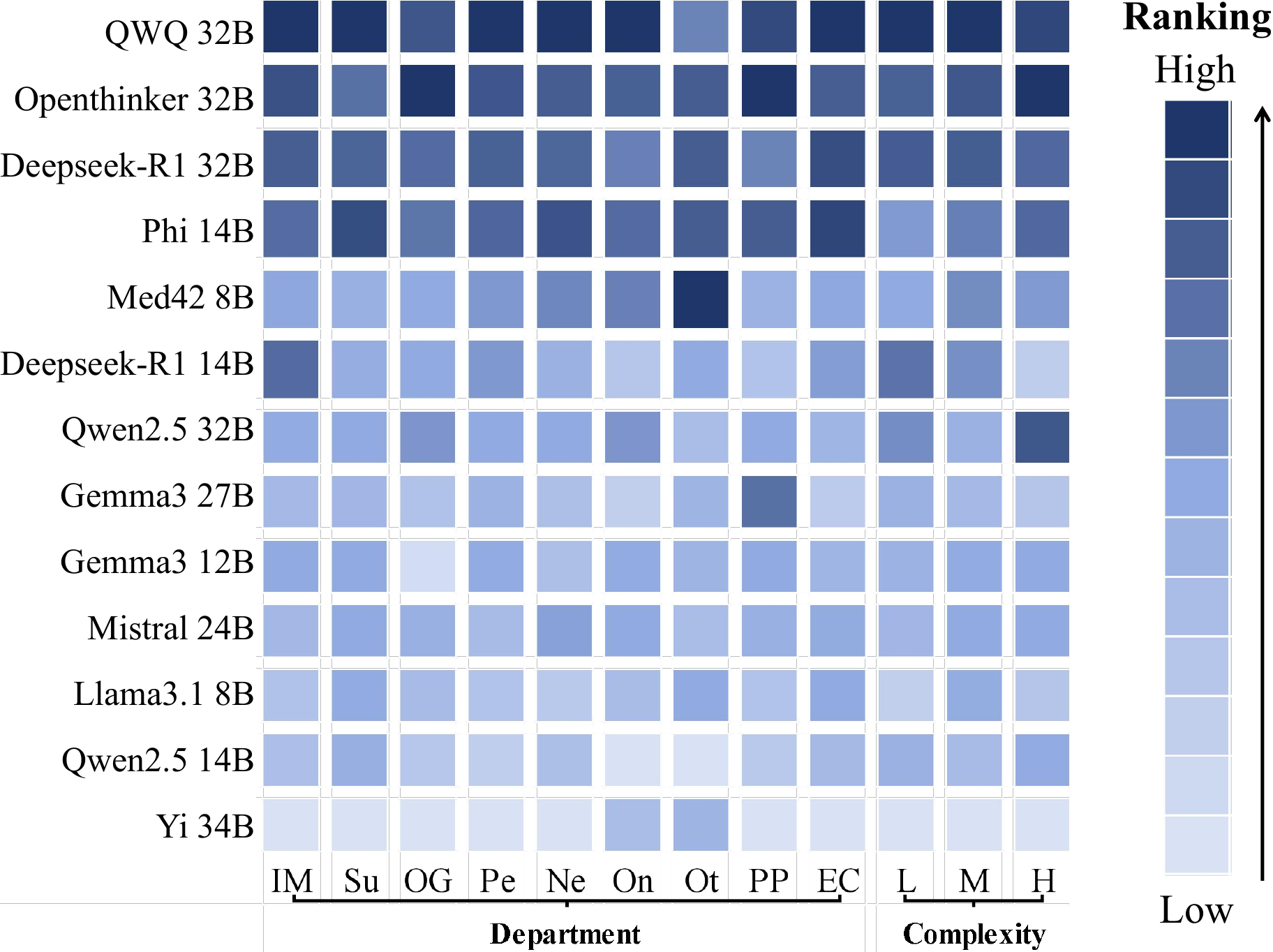}
\caption{{The visualization of the LLM expertise table for the pool of LLMs.}}
\label{fig_captable}
\end{figure}

The above classification and answers accuracy are then utilized to construct the LLM expertise table (visualized in Fig.~\ref{fig_captable}), which forms a foundation for subsequent expertise-driven agent selection. Notably, due to the establishment of standardized classification criteria, the LLM expertise table, created from a single validation set, can be applied across various test sets in the medical domain.

\subsubsection{Expertise-driven Agent Selection}

Building upon the construction of the LLM expertise table, we subsequently transition to the inference phase, which consists of two principal stages: query classification and LLM selection.

To ensure precise classification of medical queries, we first designate a dedicated query classifier from the pre-established LLM expertise table, where that classifier is selected based on its overall classification proficiency in the categories of medical departments and the difficulty levels of the queries. Specifically, for each LLM, we compute a composite classification score by the sum of its classification accuracies in medical department categories and query difficulty levels. Afterward, the LLM with the highest composite score is appointed as the \textit{query classifier}. The formal definition of this process is as follows:
\begin{equation}
\left \{\begin{array}{l}
S_n^{\text{class}} = \text{Acc}_{\text{dept}}^{n} +  \text{Acc}_{\text{diff}}^{n} \\
\text{Classifer} = \left\{ \text{LLM}_n \mid S_n^{\text{class}} \in \text{Top-}1(S^{\text{class}})\right\}
\end{array}.
\right.
\end{equation}

In the LLM selection stage, the query classifier is subsequently applied to assign a predicted department category and difficulty level to each query, where predicted classification results are served as keys to retrieve corresponding answer accuracies from the LLM expertise table. For each candidate LLM in the pool, we calculate a category-specific expertise score $S_n^{\text{ans}}$ through a weighted combination of its answer accuracies along the two dimensions. This step can be expressed as:
\begin{equation}
S_n^{\text{ans}} = \beta \cdot \text{ACC}_{\text{dept}}^{n,di} + (1 - \beta) \cdot \text{ACC}_{\text{diff}}^{n,li},
\end{equation}
where $di$ and $li$ mean the medical department category and the query difficulty levels of the current query, respectively. $\beta$ denotes the trade-off hyperparameter that is set to $0.7$ in this work. Notably, this setting aligns with the empirical observation that the difficulty-level classification, due to its coarser granularity and limited number of categories, tends to be more sensitive to LLM's global performance and thus is assigned a relatively lower weight in the composite expertise score.

Finally, based on the ranking of these expertise scores, we select the Top-$N$ LLMs to act as the specific-domain medical expert agents and form a collaborative agent set responsible for answering the query. The selection process is defined as:
\begin{equation}
\mathcal{M}_{q}^{i} = \left\{ n \mid S_n^{\text{ans}} \in \text{Top-}N(S^{\text{ans}}),\ n \in \mathcal{M} \right\},
\end{equation}
where $\mathcal{M}_{q}^{i}$ denotes the set of recruited LLMs assigned to answer the query $Q_{i}^{t}$.

\subsection{Confidence- and Adversarial-Driven Multi-agent Collaboration}
The output of a single LLM is often susceptible to the phenomenon of ``hallucination" \cite{martino2023knowledge,huang2025survey}, where the information generated from the LLM appears plausible but is factually incorrect, unsubstantiated, or misleading, significantly degrading overall performance in answering medical queries. To address this challenge, recent studies \cite{shi2025mitigating,feng2024don,kim2024mdagents,wang2025mixtureofagents} have explored the multi-agent collaboration mechanisms to leverage the complementary strengths of multiple LLMs to improve answer accuracies. However, low-quality responses from some agents can introduce noise during the collaboration process and even mislead other agents, ultimately undermining collective performances. To mitigate this issue, we propose a confidence- and adversarial-driven multi-agent collaboration strategy within a multi-layer collaboration architecture, where information from one layer is iteratively refined in subsequent layers to progressively enhance answer quality, aimed at enhancing the reliability of multi-agent collaboration in MDM.

Specifically, for each query, each recruited agent is first assigned to independently generate an initial response $R^{(1)}_n$ along with a self-assessed confidence score reflecting its belief in the correctness of its response. The collaboration process then proceeds in two parallel stages. First, in the confidence fusion stage, the self-assessed confidence score of each agent is integrated with its historical performance on the query category (\emph{i.e.,} the category-specific expertise score $S_n^{\text{ans}}$). This yields an overall confidence score $Conf^{(1)}$ for each answer, integrating both instance-level subjective certainty and category-level statistical reliability. 
In the adversarial verification stage, the agent with the highest category-specific expertise score is designated the \textit{Judge} role, which performs cross-verification on all candidate responses by assessing their factual inconsistency and identifying potential errors or contradictions. The purpose of this process is to generate informative feedback that can guide the refinement in subsequent iterations, thereby correcting misinformation to enhance overall response quality.

The outputs of the two stages, including the overall confidence score of each agent, the potential errors identified by the \textit{Judge}, and all initial responses of agents, are propagated to the next collaboration layer. This initiates a second round of response generation, where agents refine their outputs based on the integrated feedback. This process can be formalized as follows:
\begin{equation}
R_i^{(2)} = \text{Agent}_i\left(\sum_{n=1}^{N}(R_n^{(1)},Conf_n^{(1)},Err_n^{(1)})\right),
\end{equation}
where $\text{Agent}_i$ represents the $i^{th}$ recruited agent.

Finally, the responses refined through a multi-layer interaction and optimization are passed to a designated LLM serving as the \textit{Aggregator} to conduct information aggregation. It integrates all received information and generates a unified final answer $Ans_{final}$, where the process can be expressed as:
\begin{equation}
Ans_{final} = \text{Agg}\left(\sum_{n=1}^{N}(Ans_n^{(2)},Conf_n^{(2)},Err_n^{(2)})\right),
\end{equation}
where $\text{Agg}$ represents the \textit{Aggregator}.

\section{Experiments}
\subsection{Datasets and Implementation Details}
\label{Datasets and Implementation Details}
\noindent We have evaluated our EMRC framework on three publicly available medical datasets: MedQA \cite{jin2021disease}, NEJMQA \cite{katz2024gpt}, and MMLU-Pro-health (MMLUP-H) \cite{wang2024mmlu}. 
\\\textbf{The MedQA dataset} is a specialized question answering benchmark designed to evaluate and advance the capabilities of artificial intelligence models in MDM. It consists of a large collection of queries sourced from real-world medical examinations, as well as queries reflecting clinical practice. It covers a broad spectrum of medical topics, such as diagnosis, treatment, and pharmacological information. In this paper, the 1272 queries and answers from the validation set are utilized to construct the LLM expertise table during the preprocessing stage, and the 1273 queries and answers from the test set are used for evaluation.
\\\textbf{The NEJMQA dataset} is based on Israel’s 2022 national medical specialty licensing examination, encompassing five fundamental clinical domains: General Surgery, Internal Medicine, Obstetrics and Gynecology, Pediatrics, and Psychiatry. It includes a total of 655 queries in both single- and multiple-choice formats. Importantly, achieving at least a 65\% accuracy rate in each subject is a prerequisite for board certification. This passing criterion is adopted as a reference point for determining whether LLMs exhibit decision-making proficiency at the physician level.
\\\textbf{The MMLUP-H dataset} is a specialized subset within the MMLU-Pro benchmark, comprising 818 rigorously curated items spanning eight medical subfields: Virology, Professional Medicine, Nutrition, Medical Genetics, Human Aging, College-level Medicine, Clinical Knowledge, and Anatomy. Each question has undergone a comprehensive development pipeline, including initial screening, content integration, distractor enhancement, and expert validation, to maximize reasoning difficulty and ensure the reliability of healthcare-related assessments.

\textbf{Baselines.} We compare against three categories of baselines.
\textbf{1) Open-access LLMs:} This category has 16 open-access LLMs including  Medllama3 8B, HuatuoGPT-o1 8B \cite{zhang2023huatuogpt}, Med42 8B \cite{christophe2024med42}, Phi4 14B \cite{abdin2024phi}, Qwen2.5 14B, Qwen2.5 32B \cite{qwen2.5}, QWQ 32B \cite{qwq32b}, Openthinker 32B \cite{openthoughts}, Deepseek-R1 32B \cite{guo2025deepseek}, Llama3 instruct 70B \cite{meta2024introducing}, Qwen1.5 72B, Qwen1.5 110B \cite{bai2023qwen}, Qwen2.5 72B \cite{team2024qwen2}, dbrx-instruct 132B \cite{team2024introducing}, Mixtral 8x22 141B \cite{jiang2024mixtral}, and WizardLM 8x22 141B \cite{xu2023wizardlm}.
\textbf{2) Close-source LLMs:} This category has two close-source LLMs (\emph{i.e.,} GPT-4o-mini-07-18 and GPT-4-0613 \cite{achiam2023gpt}).
\textbf{3) Multi-agent collaboration:} To evaluate the effectiveness of the EMRC framework in terms of both its agent recruitment strategy and the proposed agent collaboration, we further compare it against alternative multi-agent collaboration approaches. This category has three SOTA multi-agent collaboration frameworks (\emph{i.e.,} Debate \cite{du2023improving}, MoA \cite{wang2025mixtureofagents}, and SelfMoA \cite{li2025rethinking}).

\begin{figure*}[htbp]
    \centering
    \includegraphics[width=0.80\textwidth]{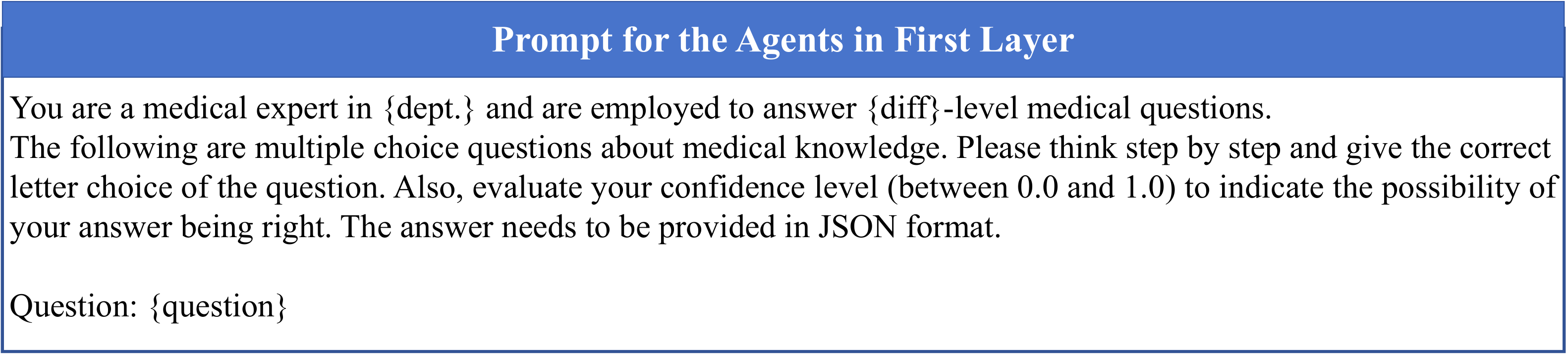}

    \vspace{0.5em}

    \includegraphics[width=0.80\textwidth]{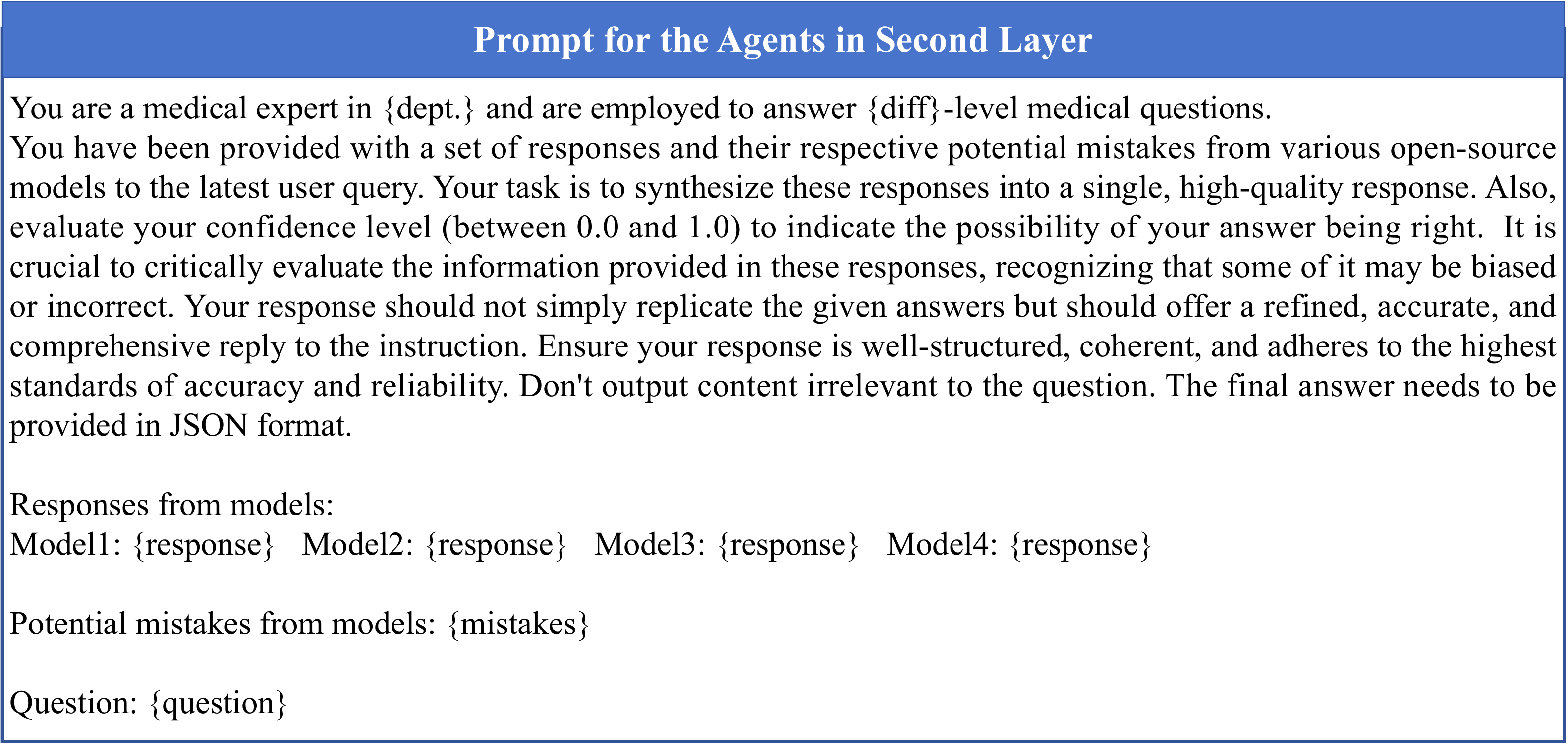}

    \vspace{0.5em}

    \includegraphics[width=0.80\textwidth]{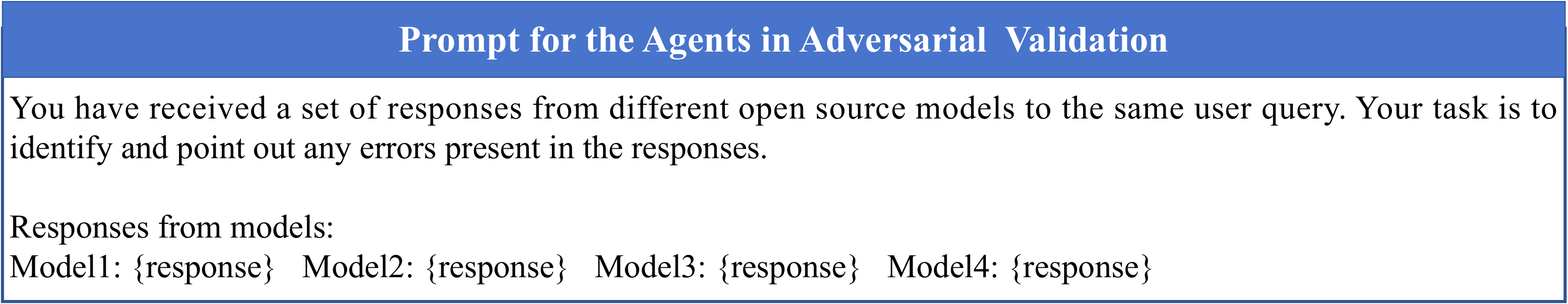}

    \vspace{0.5em}

    \includegraphics[width=0.80\textwidth]{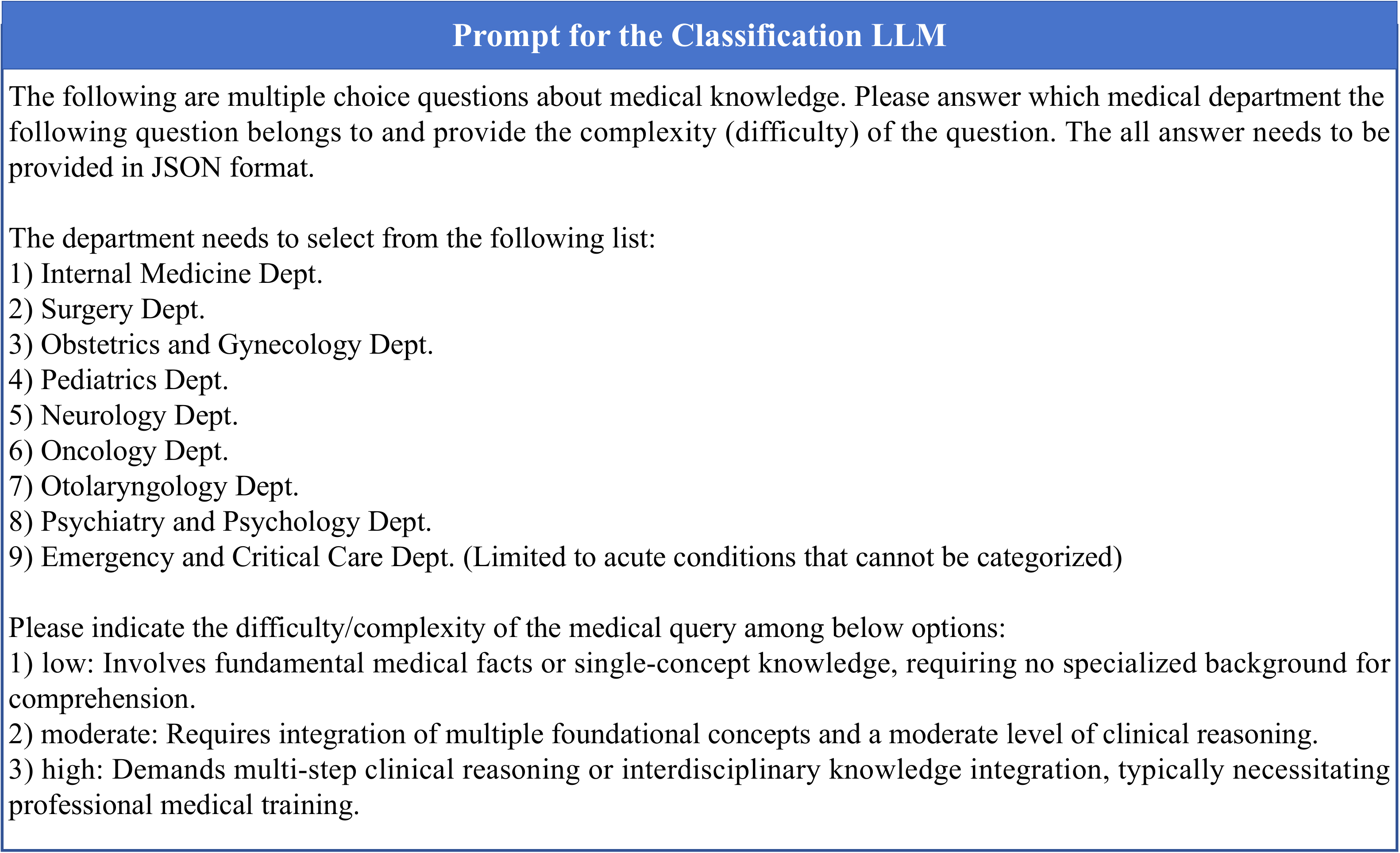}

    \caption{Prompts in our EMRC framework.}
    \label{fig_prompt}
\end{figure*}

\textbf{Implementation Details.} To balance computational efficiency with strong performance, we exclusively utilize open-access LLMs ranging from 8B to 34B parameters to construct the pool of LLMs, including Med42 8B \cite{christophe2024med42}, Llama3.1 8B \cite{grattafiori2024llama}, Gemma3 12B \cite{team2025gemma}, Phi4 14B \cite{abdin2024phi}, QWen2.5 14B \cite{qwen2.5}, Deepseek-R1 14B \cite{guo2025deepseek}, mistral-small 24B, Gemma3 27B \cite{team2025gemma}, QWen2.5 32B \cite{qwen2.5}, Deepseek-R1 32B \cite{guo2025deepseek}, QWQ 32B \cite{qwq32b}, Openthinker 32B\cite{openthoughts}, and Yi 34B \cite{young2024yi}. 
A single 34B-parameter model requires approximately 22 GB of GPU memory, which fits comfortably within the 24GB capacity of a NVIDIA GeForce RTX 4090, making this setup technically and economically viable. The prompt of our framework is shown in Fig.~\ref{fig_prompt}. All implementations are carried out using the PyTorch framework on NVIDIA GeForce RTX 4090. Throughout the study, we strictly adhere to the licensing terms associated with each model employed. 

\textbf{Metrics.} To comprehensively evaluate the performance of the compared models and our EMRC framework, we exploit a series of evaluation metrics, including ACCuracy (ACC), weighted F1-score (F1), PREcision (PRE) \cite{powers2020evaluation}, SPEcificity (SPE) \cite{saah1998sensitivity}, Matthews Correlation Coefficient (MCC) \cite{matthews1975comparison}, and Cohen's Kappa (CK) \cite{mchugh2012interrater}.

\begin{table*}[htbp]
  \centering
  \renewcommand{\arraystretch}{1.1}
  \renewcommand{\tabcolsep}{0.4mm}
  \caption{Comparison between our EMRC framework and other SOTA methods in terms of ACC on MMLUP-H dataset. The \textbf{best}/\underline{second best} results are marked in bold/underline.}
    \begin{tabular}{l|cccccccc|c}
    \toprule
    \multirow{2}[2]{*}{Methods} & \multicolumn{9}{c}{MMLUP-H} \\
          \cmidrule{2-10}          & \multicolumn{1}{>{\centering}m{5em}}{Virology} & \multicolumn{1}{>{\centering}m{5em}}{Professional Medicine} & \multicolumn{1}{>{\centering}m{5em}}{Nutrition} & \multicolumn{1}{>{\centering}m{5em}}{Medical Genetics} & \multicolumn{1}{>{\centering}m{4em}}{Human Aging} & \multicolumn{1}{>{\centering}m{5em}}{College Medicine} & \multicolumn{1}{>{\centering}m{5em}}{Clinical Knowledge} & \multicolumn{1}{>{\centering}m{5em}}{Anatomy} & \multicolumn{1}{>{\centering}m{5em}}{Acc.} \\
    \midrule
    \rowcolor[rgb]{ 1,  .949,  .8}\multicolumn{10}{c}{Open-access LLMs} \\
    Medllama3 8B & 36.96  & 53.94  & 48.60  & 53.70  & 33.72  & 54.16  & 51.39  & 34.18  & 47.43  \\
    HuatuoGPT-o1 8B & 60.87  & 60.24  & 54.75  & 51.85  & 41.86  & 68.75  &  55.56 &  48.10 &  55.99 \\
    Med42 8B & 60.87  & 46.06  & 43.02  & 44.44  & 31.40  & 52.08  &  43.06 &  34.18 &  43.15 \\
    Phi4 14B & 52.17  & 72.83  & 68.16  & 83.33  & 45.35  & \underline{85.42}  & 73.61  & 65.82  & 68.58  \\
    Qwen2.5 14B & 63.04  & 62.20  & 63.13  & 74.07  & 45.35  & 83.33  & 65.28  & 54.43  & 62.22  \\
    Qwen2.5 32B & \underline{65.22}  & 70.08  & 69.27  & 75.93  & 52.33  & 79.17  & 73.61  & 59.49  & 67.97  \\
    QWQ 32B & 58.70  & 68.50  & 67.04  & 68.52  & \underline{54.65}  & 75.00  & 72.22  & 63.29  & 66.38  \\
    Openthinker 32B & 63.04  & 70.47  & 64.25  & 77.78  & 51.16  & 79.17  & \textbf{75.00}  & 67.09  & 67.73  \\
    Deepseek-R1 32B & 60.87  & 57.87  & 58.66  & 70.37  & 50.00  & 77.08  & 61.11  & 54.43  & 59.29  \\
    Llama3 instruct 70B & \underline{65.22}  & 72.44  & 69.27  & 77.78  & 50.00  & 72.92  & 63.89  & 64.56  & 67.85  \\
    Qwen1.5 72B & 23.91  & 10.24  & 22.35  & 14.81  & 12.79  & 10.42  & 9.72  & 16.46  & 14.79  \\
    Qwen1.5 110B & 58.70  & 45.67  & 49.72  & 51.85  & 39.53  & 56.25  & 48.61  & 49.37  & 48.29  \\
    Qwen2.5 72B & 60.87  & 72.83  & 68.72  & 75.93  & 51.16  & 77.08  & 68.06  & \underline{68.35}  & 68.58  \\
    dbrx-instruct 132B & 41.30  & 41.73  & 45.81  & 57.41  & 32.56  & 45.83  & 31.94  & 39.24  & 41.81  \\
    Mixtral 8x22 141B & 65.22  & 49.61  & 58.10  & 70.37  & 41.86  & 70.83  & 62.50  & 50.63  & 55.38  \\
    WizardLM 8x22 141B & 50.00  & 40.94  & 55.31  & 62.96  & 53.49  & 64.58  & 52.78  & 48.10  & 50.49  \\
    \midrule
    \rowcolor[rgb]{ .89,  .949,  .851} \multicolumn{10}{c}{Close-source LLMs} \\
    GPT-4o-mini-07-18 & 60.87  & 72.83  & 65.36  & 74.07  & 51.16  & 77.08  & \textbf{75.00}  & 58.23  & 67.36  \\
    GPT-4-0613 & 60.87  & \textbf{78.35}  & \underline{70.39}  & \underline{81.48}  & \underline{54.65}  & 81.25  & \textbf{75.00}  & 63.29  & \underline{71.76}  \\
    \midrule
    \rowcolor[rgb]{ .824,  .957,  .949} \multicolumn{10}{c}{ Multi-agent Collaboration} \\
    Debate & 58.70  & 72.44  & 68.16  & 75.93  & 47.67  & 85.42  & \textbf{75.00}  & 67.09  & 68.83  \\
    MoA   & 63.04  & 54.72  & 58.66  & 62.96  & 53.49  & 64.58  & 52.78  & 55.70  & 56.97  \\
    SelfMoA & 34.78  & 47.24  & 44.13  & 62.96  & 38.37  & 54.17  & 54.17  & 49.37  & 47.19  \\
    \midrule
    \midrule
    \textbf{Ours} & \textbf{67.39} & \underline{77.95} & \textbf{73.74} & \textbf{87.04} & \textbf{58.14} & \textbf{89.58} & \textbf{75.00} & \textbf{68.35} & \textbf{74.45} \\
    \bottomrule
    \end{tabular}%
  \label{tab:compare_MP}%
\end{table*}%

\begin{table*}[htbp]
  \centering
  \caption{Comparison between our EMRC framework and other SOTA methods in terms of ACC on NEJMQA and MedQA datasets. The \textbf{best}/\underline{second best} results are marked in bold/underline.}
    \begin{tabular}{l|ccccc|c||c}
    \toprule
    \multirow{2}[4]{*}{Methods} & \multicolumn{6}{c|}{NEJMQA} & {MedQA} \\
\cmidrule{2-8}          & \multicolumn{1}{>{\centering}m{5em}}{General Surgery} & \multicolumn{1}{>{\centering}m{5em}}{Internal Medicine} & \multicolumn{1}{>{\centering}m{5em}}{Obstetrics Gynecology} & \multicolumn{1}{>{\centering}m{5em}}{Pediatrics} & \multicolumn{1}{>{\centering}m{5em}}{Psychiatry} & \multicolumn{1}{>{\centering}m{5em}|}{Acc.} & \multicolumn{1}{>{\centering}m{5em}}{Acc.} \\
    \midrule
    \rowcolor[rgb]{ 1,  .949,  .8} \multicolumn{8}{c}{Open-access LLMs} \\
    Medllama3 8B & 30.50  & 32.54  &  27.34 &  38.38 & 50.00  &  37.86 & 49.49\\
    HuatuoGPT-o1 8B &  46.81 & 43.65  & 36.69  & 59.60  & 63.33  &  49.77 & 62.45\\
    Med42 8B &  41.84 & 37.30  & 41.73  & 46.46  & 60.00  &  46.11 & 59.15\\
    Phi4 14B & 43.26  & 43.65  & 35.97  & 49.49  & 49.33  & 44.12 & 74.86\\
    Qwen2.5 14B & 43.97  & 51.59  & 43.17  & 52.53  & 62.67  & 50.84 & 63.16\\
    Qwen2.5 32B & 52.48  & 61.11  & 53.24  & 59.60  & 68.67  & 59.08 & 68.58 \\
    QWQ 32B & 62.41  & 65.08  & 46.04  & 66.67  & 76.00  & 63.21 & 78.63\\
    Openthinker 32B & 60.99  & 65.08  & 51.80  & \underline{71.72}  & 74.00  & 64.43 & 79.18\\
    Deepseek-R1 32B & 60.99  & 62.70  & 51.80  & 61.62  & 65.33  & 60.46 & 77.30\\
    Llama3 instruct 70B & 60.28  & 56.35  & 57.55  & 67.68  & 69.33  & 62.14 & 72.74\\
    Qwen1.5 72B & 34.75  & 38.89  & 37.41  & 42.42  & 48.67  & 40.46 & 50.75 \\
    Qwen1.5 110B & 52.48  & 47.62  & 43.88  & 50.51  & 70.00  & 53.44 & 61.98\\
    Qwen2.5 72B & 58.16  &  59.52 &  53.96 &  59.60 &  70.00 & 60.46 & 74.71\\
    dbrx-instruct 132B & 47.52  & 43.65  & 38.85  & 45.45  & 51.33  & 45.50 & 42.73\\
    Mixtral 8x22 141B & 53.90  & 48.41  & 52.52  & 54.55  & 68.67  & 56.03 & 62.06 \\
    WizardLM 8x22 141B & 49.65  & 48.41  & 46.76  & 58.59  & 68.00  & 54.35 & 72.82 \\
    \midrule
    \rowcolor[rgb]{  .89,  .949,  .851} \multicolumn{8}{c}{Close-source LLMs} \\
    GPT-4o-mini-07-18 & 52.48  & 57.14  & 41.01  & 61.62  & 74.00  & 57.25 & 69.05\\
    GPT-4-0613 & 62.41  & 65.87  & 56.12  & 69.70  & \underline{78.00}  & 66.41 & 73.61\\
    \midrule
    \rowcolor[rgb]{ .824,  .957,  .949} \multicolumn{8}{c}{ Multi-agent Collaboration} \\
    Debate & \underline{64.54}  & \underline{69.05}  & \underline{58.27}  & \underline{71.72}  & 76.67  & \underline{67.94} & \underline{79.26} \\
    MoA   & 51.77  & 43.65  & 48.92  & 55.56  & 70.00  & 54.35 & 58.44\\
    SelfMoA & 39.01  & 27.78  & 33.81  & 40.40  & 53.33  & 39.24 & 61.74\\
    \midrule
    \midrule
    \textbf{Our} & \textbf{66.67} & \textbf{77.78} & \textbf{58.99} & \textbf{74.75} & \textbf{82.00} & \textbf{71.91} & \textbf{83.42}\\
    \bottomrule
    \end{tabular}%
  \label{tab:compare_Nej}%
\end{table*}%

\begin{figure*}[!t]
\centering
\includegraphics[width=0.99\textwidth]{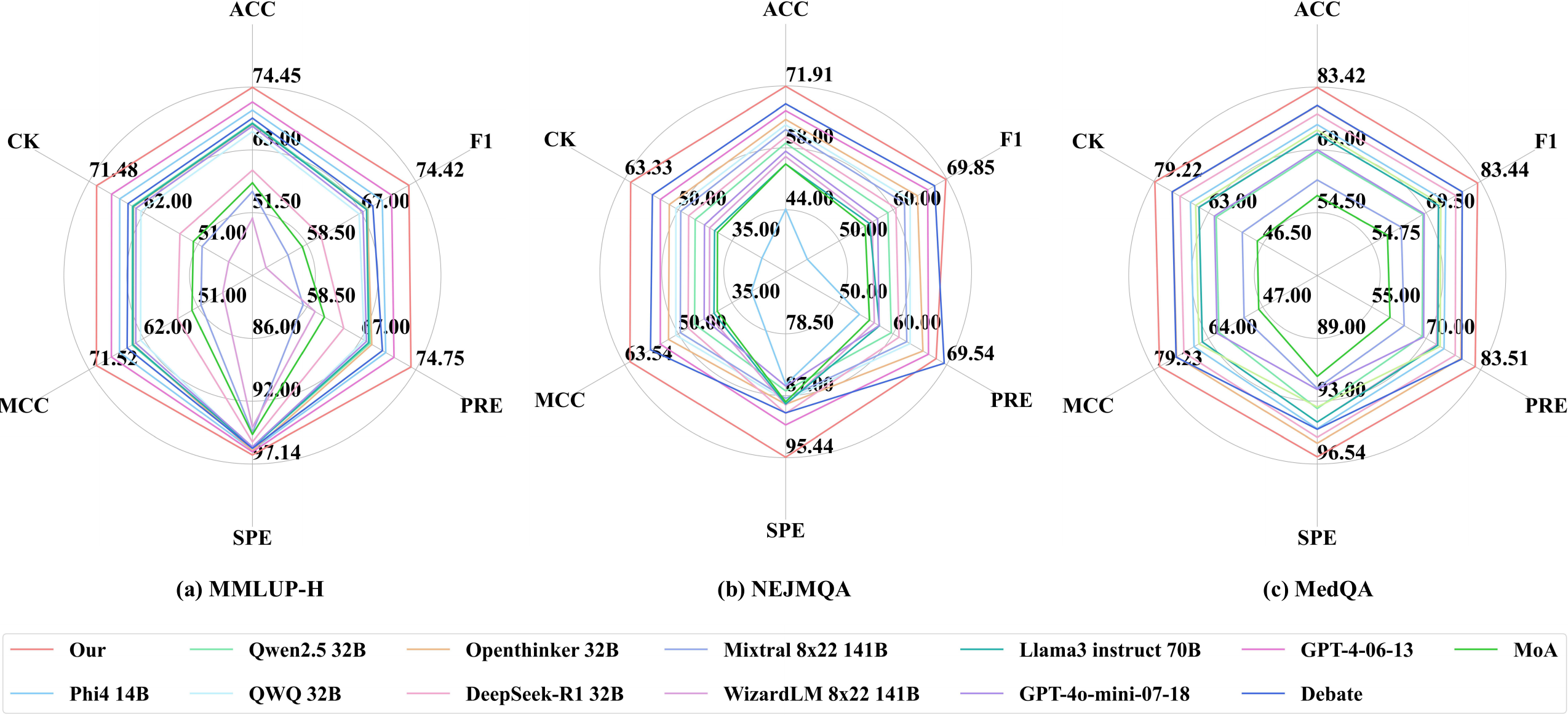}
\caption{{Comparison between our EMRC framework and other representative SOTA methods in terms of ACC, F1, PRE, SPE, MCC, and CK metrics on three datasets.}}
\label{fig_metrics}
\end{figure*}

\subsection{Comparison with the State-of-the-art Methods}
\label{Comparison with the State-of-the-arts}

To verify the effectiveness of our EMRC framework, we compare our method with 21 baselines in the MMLUP-H, NEJMQA, and MedQA datasets. The results are shown in Tables~\ref{tab:compare_MP},~\ref{tab:compare_Nej}, and Fig.~\ref{fig_metrics}. From the results, we can observe that our EMRC framework outperforms all comparison methods in terms of all datasets. Specifically, on the MMLUP-H dataset, EMRC achieves an average accuracy of 74.45\%, surpassing the best single LLM, GPT-4-0613, which scored 71.76\%, and other collaborative frameworks such as Debate, which scored 68.83\%. 
On the NEJMQA dataset, our EMRC attained an average accuracy of 71.91\%, outperforming the best individual LLM, GPT-4-0613, with 66.41\%, and Debate with 67.94\%. Additionally, EMRC achieved an accuracy of 83.42\% on the MedQA dataset, also outperforming all other comparison methods. Meanwhile, we present additional evaluation metrics in Fig.~\ref{fig_metrics}. From the results, we can find that our EMRC framework encloses the largest area across all radar charts, providing multi-dimensional evidence of its superior overall performance.

Notably, our framework consistently achieves superior performance across almost all subcategories on NEJMQA and MMLUP-H datasets. This is due to the effective combination of expertise-aware agent recruitment and complementary multi-agent collaboration. By dynamically selecting the most appropriate LLMs for each medical query, based on their specialized expertise in different medical departments and varying levels of query difficulty, the framework ensures that each LLM contributes its strongest capabilities to the decision-making process. The construction of the LLM expertise table plays a crucial role in this process, as it allows the framework to evaluate the strengths of each model across various domains and recruit the most suitable agents. Moreover, the confidence- and adversarial-driven multi-agent collaboration enhances response accuracy by integrating outputs from multiple agents while cross-validating for inconsistencies or errors. This collaborative approach ensures that only reliable information contributes to the final diagnosis. The synergy between expertise-specific agent recruitment and multi-agent collaboration leads to significant performance improvements across all medical subcategories, enabling the framework to outperform existing state-of-the-art methods and deliver superior diagnostic accuracy and reliable decision support.

\subsection{Ablation Studies}
\label{Ablation Studies}
In this section, we will conduct extensive experiments to validate the effectiveness of the key components of our EMRC framework on the MMLUP-H and NEJMQA datasets. 

{\bf 1) Effectiveness of expertise-aware agent recruitment strategy.}
Our expertise-aware agent recruitment strategy dynamically selects LLMs based on their expertise-specific knowledge and proficiency to tailor agent teams for each medical query. To validate the effectiveness of the proposed recruitment strategy, we design four comparative variants. In the first two variants, multiple LLMs are randomly recruited for each query to serve as collaborative agents. In the latter two variants, we recruit the top-N LLMs based on their overall task performance, specifically those with the highest average accuracy on the validation set.
Experimental results are presented in Table~\ref{tab:abf1}, where the proposed expertise-aware agent recruitment strategy consistently outperforms all alternative approaches. In contrast, the random selection strategy leads to the most substantial performance degradation, likely due to the high probability of recruiting weaker models, which negatively impacts overall answer quality. While selecting top-performing models based on global accuracy may seem intuitive, its performance still lags behind our method.
These findings suggest that relying solely on overall task performance may not adequately reflect a model’s suitability for specific categories of medical queries. In comparison, our expertise-aware agent recruitment strategy offers a more adaptive approach to model selection, demonstrating superior effectiveness in multi-agent collaboration settings.



\begin{table}[t]
\centering
  \renewcommand{\arraystretch}{1.1}
  \renewcommand{\tabcolsep}{1mm}
\caption{Ablation studies of EMRC on MMLUP-H and NEJMQA datasets for recruitment and collaboration strategies. Best results in each block are in \textbf{bold}.}
\label{tab:ablation_recruitment_collab}
\begin{tabular}{l l c c}
\toprule
\textbf{Setting} & \textbf{Variant} & \textbf{MMLUP-H} & \textbf{NEJMQA} \\
\midrule
\multirow{5}{*}{\makecell[l]{\textbf{Agent Recruitment}\\\textbf{Strategy}}} 
& Random-3 & 60.64 & 54.05 \\
& Random-4 & 62.59 & 56.18 \\
& Task-level Top-3 & 70.42 & 66.56 \\
& Task-level Top-4 & 69.44 & 67.94 \\
& \textbf{Expertise-aware (Ours)} & \textbf{74.45} & \textbf{71.91} \\
\midrule
\multirow{4}{*}{\makecell[l]{\textbf{Collaboration}\\\textbf{Strategy}}} 
& Baseline (no fusion/adv.) & 70.29 & 67.94 \\
& w/o Confidence & 72.13 & 69.16 \\
& w/o Adversarial & 72.86 & 70.23 \\
& \textbf{Full (Ours)} & \textbf{74.45} & \textbf{71.91} \\
\bottomrule
\end{tabular}
\label{tab:abf1}%
\end{table}

{\bf 2) Effectiveness of confidence- and adversarial-driven multi-agent collaboration strategy.}
The confidence- and adversarial-driven multi-agent collaboration strategy integrates LLM responses using the confidence fusion and adversarial verification, where a highly expert LLM cross-validates outputs to enhance the diagnostic reliability and produce a unified answer via a multi-layer agent collaboration network. 
To evaluate the effectiveness of the proposed architecture, we design three ablation variants. In the first variant, \emph{i.e.,} Baseline, both the confidence fusion and adversarial verification components are removed, retaining only simple interactions among model outputs. In the second variant, \emph{i.e.,} w/o Confidence, the confidence fusion process is excluded, where only the self-assessed confidence score of each agent are utilized during inference. In the third variant, \emph{i.e.,} w/o Adversarial, the adversarial verification stage is removed.
The experimental results, as presented in Table~\ref{tab:abf1}, clearly indicate that the full model, which incorporates both the confidence fusion and adversarial verification mechanisms, delivers superior performance across all evaluation metrics. Notably, this comprehensive model significantly outperforms the ablated variants, demonstrating its robustness. These findings offer compelling evidence of the effectiveness of our multi-agent collaboration mechanism, which substantially enhances both the integration of diverse information and the accuracy of the answers generated in a multi-agent LLM environment. This success reinforces the critical role of these mechanisms in improving the reliability of medical decision-making systems, particularly when leveraging complex, domain-specific knowledge from multiple sources.


\begin{table}[t]
\centering
\renewcommand{\arraystretch}{1.1}
\renewcommand{\tabcolsep}{1.5mm}
\caption{Ablation studies of EMRC on MMLUP-H and NEJMQA datasets for the number of recruited agents and collaboration layers. Best results in each block are in \textbf{bold}.}
\label{tab:ablation_agents_layers}
\begin{tabular}{l l c c}
\toprule
\textbf{Setting} & \textbf{Variant} & \textbf{MMLUP-H} & \textbf{NEJMQA} \\
\midrule
\multirow{5}{*}{\makecell[l]{\textbf{Number of}\\\textbf{Recruited Agents}}} 
& 1 & 70.78 & 67.63 \\
& 2 & 71.88 & 69.31 \\
& 3 & 73.47 & 71.15 \\
& \textbf{4 (Ours)} & \textbf{74.45} & \textbf{71.91} \\
& 5 & 73.10 & 71.75 \\
\midrule
\multirow{3}{*}{\makecell[l]{\textbf{Number of}\\\textbf{Collaboration Layers}}} 
& 1 & 72.13 & 70.68 \\
& \textbf{2 (Ours)} & \textbf{74.45} & \textbf{71.91} \\
& 3 & 72.98 & 69.92 \\
\bottomrule
\end{tabular}
\label{tab:abf2}%
\end{table}

{\bf 3) Impact on the number of recruited agents.}
We investigate the influence of the number of recruited agents on framework performance, with the experimental results summarized in Table~\ref{tab:abf2}. The findings indicate a consistent increase in overall answer accuracy as the number of agents grows, demonstrating the beneficial effect of multi-agent collaboration in enhancing response accuracy. This trend reflects the added value of incorporating a diverse range of expertise from multiple agents, which collectively improves the reliability and precision of the output.
However, when the number of recruited agents reaches five, a slight decline in performance is observed. This suggests that expanding the agent pool beyond a certain threshold can introduce models with comparatively weaker capabilities, which may generate lower-quality responses. These suboptimal contributions can undermine the final synthesis of the answers, leading to a marginal degradation in overall performance. The phenomenon highlights a critical insight: while diversity in model selection is beneficial, it must be balanced with the quality of the agents involved. Therefore, in multi-agent collaboration systems, it is essential to control the number of recruited agents to ensure that the benefits of model diversity are not outweighed by the potential negative effects of incorporating less capable models.



\begin{figure*}
\centering
\includegraphics[width=0.80\textwidth]{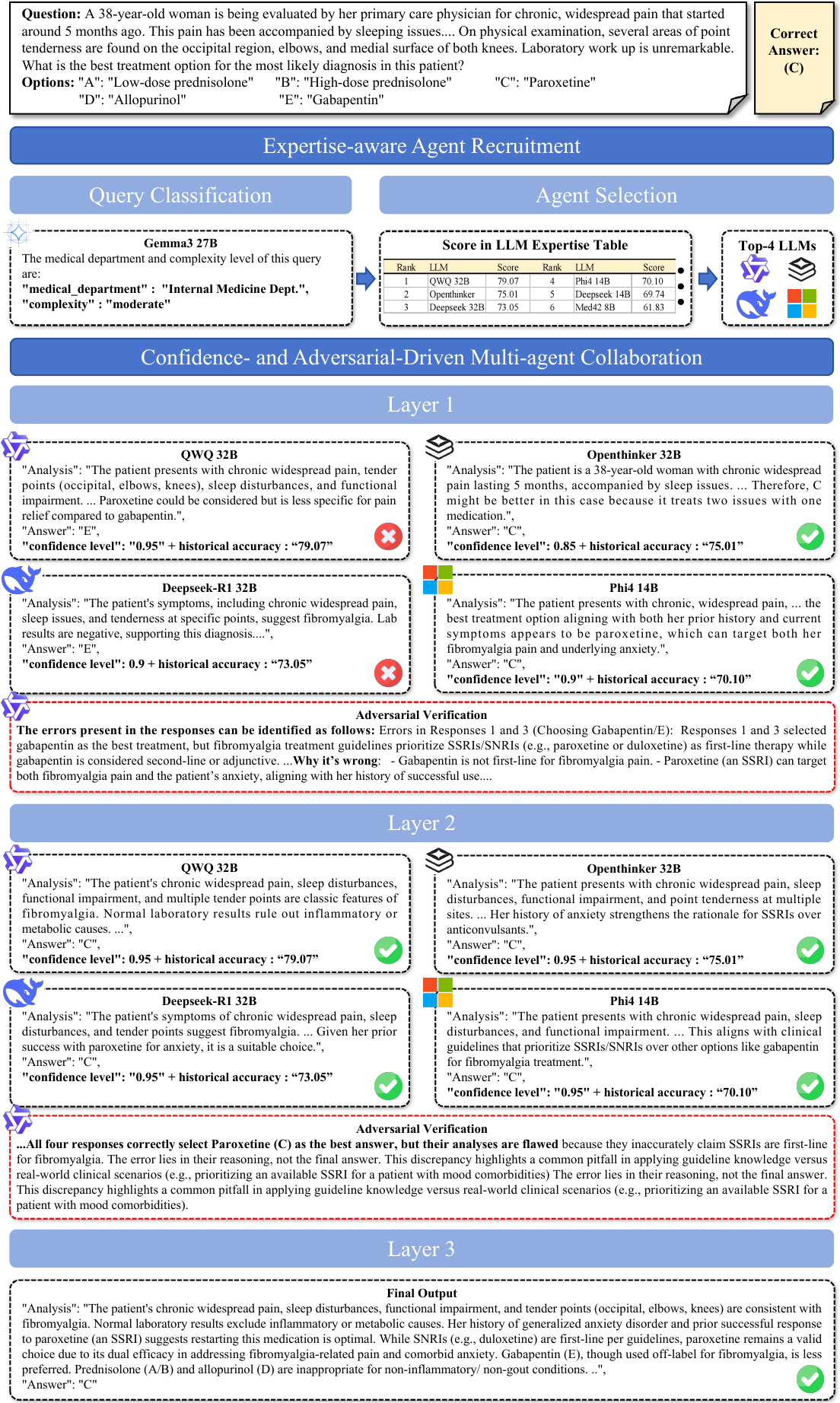}
\caption{{The case study in the MedQA dataset.}}
\label{fig_case_study}
\end{figure*}


{\bf 4) Impact on the number of collaboration layers.}
We further investigate the impact of the number of interaction layers on the performance of the multi-agent collaboration framework, where experimental results are presented in Table~\ref{tab:abf2}. The results show that the diagnostic accuracy improves as the number of layers increases from one to two, but begins to degrade when additional layers are introduced. Specifically, the framework achieves its highest accuracy with two collaboration layers. In contrast, a single-layer setup yields lower performance, indicating that one round of interaction is insufficient for fully refining low-quality responses and integrating adversarial feedback. However, adding a third layer leads to a performance drop. This degradation occurs because the generated outputs from the LLMs begin to diverge from the predefined format specifications, resulting in inconsistencies that complicate the subsequent stages of answer filtering and integration. This phenomenon suggests that while iterative interaction is valuable, excessive rounds of collaboration may introduce unwanted effects, such as format drift and semantic noise.

\subsection{Case Study}
In this section, we provide an example of our framework on the MedQA dataset, as shown in Fig.~\ref{fig_case_study}. 
In the expertise-aware agent recruitment stage, the EMRC framework first classifies the query using the Deepseek-R1 model (671B parameters), which identifies the query as belonging to the cardiology department with a high difficulty level due to its requirement for precise diagnostic reasoning. Based on the LLM expertise table constructed during the preprocessing phase, the framework selects four LLMs with the highest expertise scores for cardiology and high-difficulty queries. These models form the collaborative agent set for this query. In the confidence- and adversarial-driven multi-agent collaboration stage, each recruited LLM generates an independent response along with a self-assessed confidence score. The category-specific expertise scores from the LLM expertise table are combined with these self-assessed confidence scores to compute an overall confidence score for each response. The LLM with the highest expertise score in Cardiology, is designated as the \textit{Judge} and performs adversarial verification. It cross-validates the responses, confirming consistency across all three diagnoses and identifying no factual errors. The responses are then passed to the \textit{Aggregator}, which integrates the inputs and produces a unified final answer.

This case study reveals the EMRC framework's strength in leveraging complementary LLM expertise to enhance diagnostic accuracy. By dynamically selecting models with proven proficiency in cardiology and ensuring robust collaboration through confidence fusion and adversarial validation, the framework achieves a reliable and accurate diagnosis, outperforming individual LLMs and other collaborative methods. The process mirrors real-world MDM, where specialists collaborate to reach a consensus, thereby improving the accuracy and reliability of automated medical decision support systems.

\section{Conclusion}
This paper presents the Expertise-aware Multi-LLM Recruitment and Collaboration (EMRC) framework to enhance medical decision-making (MDM) by dynamically selecting and integrating multiple large language models (LLMs). The framework consists of two key stages. The first stage, expertise-aware agent recruitment, involves the evaluation and selection of LLMs based on their proficiency in classifying and answering queries across various medical departments and difficulty level of queries. The core of this stage is the construction of the LLM expertise table, which systematically catalogs the strengths of each LLM, thereby guiding the dynamic recruitment of the most appropriate and proficient LLMs for each specific query. The second stage involves confidence- and adversarial-driven multi-agent collaboration, where the responses from the recruited agents are integrated using confidence fusion and adversarial validation to refine the final decision, ensuring both accuracy and reliability. Experimental results on the MedQA, NEJMQA, and MMLU-Pro-Health datasets demonstrate that the EMRC framework consistently outperforms individual state-of-the-art LLMs and other multi-agent collaboration methods, showing substantial improvements in diagnostic accuracy. These findings highlight the effectiveness of combining expertise-driven recruitment with rigorous collaboration mechanisms, laying a solid foundation for the development of more robust, reliable, and accurate AI-assisted decision support systems in the future.

\ifCLASSOPTIONcaptionsoff
  \newpages
\fi

\bibliographystyle{IEEEtran}
\bibliography{refs}

\end{document}